\documentclass[letterpaper, 10 pt, conference]{ieeeconf}  %

\IEEEoverridecommandlockouts                              %

\usepackage{amsmath} %
\usepackage{amssymb}  %
\usepackage{hyperref}
\usepackage{booktabs}
\usepackage{multirow}
\usepackage{makecell}
\usepackage{graphicx}
\usepackage{wrapfig}
\usepackage{subcaption}
\usepackage{xcolor}
\usepackage{pgfplots}
\usetikzlibrary{patterns}
\pgfplotsset{compat=1.18}

\makeatletter
\let\NAT@parse\undefined
\makeatother
\usepackage[numbers,sort&compress]{natbib}

\definecolor{myblue}{RGB}{66,133,244}
\definecolor{myred}{RGB}{234,67,53}
\definecolor{myyellow}{RGB}{251,188,4}
\definecolor{mygreen}{RGB}{52,168,83}

\title{\LARGE \bf
RPMArt: Towards Robust Perception and Manipulation\\ for Articulated Objects
}

\author{Junbo Wang$^{1}$, Wenhai Liu$^{1}$, Qiaojun Yu$^{1}$, Yang You$^{2}$, Liu Liu$^{3}$, Weiming Wang$^{1}$ and Cewu Lu$^{1*}$%
\thanks{$^{1}$Junbo Wang, Wenhai Liu, Qiaojun Yu, Weiming Wang, Cewu Lu are with Shanghai Jiao Tong University, China. *Cewu Lu is the corresponding author. 
        Email: {\tt\small \{sjtuwjb3589635689, sjtu-wenhai, yqjllxs, wangweiming, lucewu\}@sjtu.edu.cn}}%
\thanks{$^{2}$Yang You is with Stanford University, U.S.A. 
        Email: {\tt\small yangyou@stanford.edu}}%
\thanks{$^{3}$Liu Liu is with Hefei University of Technology, China. 
        Email: {\tt\small liuliu@hfut.edu.cn}}%
}

\begin{document}

\maketitle
\thispagestyle{empty}
\pagestyle{empty}

\begin{abstract}

Articulated objects are commonly found in daily life. It is essential that robots can exhibit robust perception and manipulation skills for articulated objects in real-world robotic applications. However, existing methods for articulated objects insufficiently address noise in point clouds and struggle to bridge the gap between simulation and reality, thus limiting the practical deployment in real-world scenarios. To tackle these challenges, we propose a framework towards Robust Perception and Manipulation for Articulated Objects (RPMArt), which learns to estimate the articulation parameters and manipulate the articulation part from the noisy point cloud. Our primary contribution is a Robust Articulation Network (RoArtNet) that is able to predict both joint parameters and affordable points robustly by local feature learning and point tuple voting. Moreover, we introduce an articulation-aware classification scheme to enhance its ability for sim-to-real transfer. Finally, with the estimated affordable point and articulation joint constraint, the robot can generate robust actions to manipulate articulated objects. After learning only from synthetic data, RPMArt is able to transfer zero-shot to real-world articulated objects. Experimental results confirm our approach's effectiveness, with our framework achieving state-of-the-art performance in both noise-added simulation and real-world environments. Code, data and more results can be found on the project website at \url{https://r-pmart.github.io}.

\end{abstract}

\section{Introduction}

\begin{figure}[!htbp]
\centering
\includegraphics[width=\columnwidth]{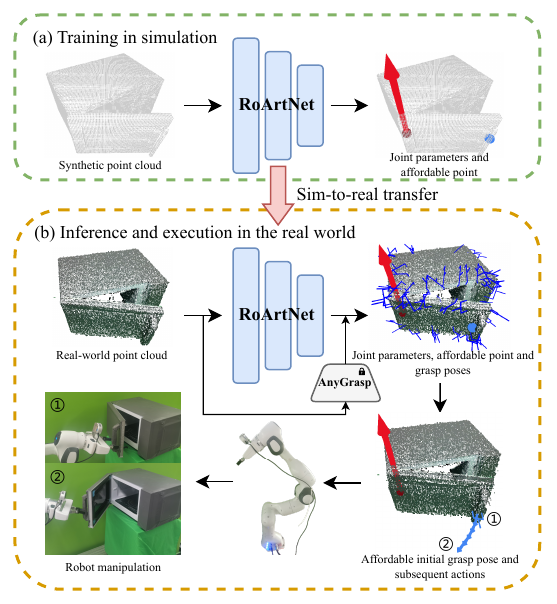}
\caption{\small RPMArt framework to tackle the real-world articulated objects perception and manipulation. (a) During training, voting targets are generated by part segmentation, joint parameters and affordable points from the simulator to supervise RoArtNet. (b) Given the real-world noisy point cloud observation, RoArtNet can still generate robust joint parameters and affordable points estimation by point tuple voting. Then, affordable initial grasp poses can be selected from AnyGrasp-generated grasp poses based on the estimated affordable points, and subsequent actions can be constrained by the estimated joint parameters.}
\label{fig:teaser}
\vspace{-5mm}
\end{figure}

Human life is populated with articulated objects, ranging from household appliances such as microwaves and refrigerators, to storage units such as safes and cabinets. Robust perception and manipulation for those objects by robots in the real world can liberate humans from mundane daily tasks. Composed of more than one rigid part connected by joints allowing rotational or translational movements, articulated objects own high degree of freedom and large state space, which makes the visual perception and downstream manipulation challenging \cite{vat-mart}. However, such geometric structure and physical constraints also provide useful clues for their perception and manipulation (see Fig. \ref{fig:teaser} (b)).

Recently, with the development of deep learning, substantial efforts have been devoted to studying the perception and manipulation for articulated objects. Prior works adapted powerful point cloud processing networks to estimate the kinematic articulation structures and parameters \cite{shape2motion, kno, ancsh, omad}, and leveraged them to produce corresponding action trajectories \cite{articulated-mm, gamma}. Another line of works explored manipulation tasks by directly imitating end-to-end demonstrations or reinforcement learning \cite{clil, maniskill2}. Despite their success, building robust and reliable robots to manipulate articulated objects within noisy observations in the real world has not yet been investigated well. To achieve this goal, two  primary challenges need to be addressed. (i) Point clouds from the real world are often noisy due to bad lighting and depth camera measurement error, while real-world articulation datasets are rare and always expensive to acquire. As a result, it is essential to introduce sim-to-real techniques to bridge the gap when training only on synthetic data. (ii) Articulated object manipulation involves both semantic and physical requirements. Grasping the relevant part requires semantic understanding of the object, and the action space is constrained by the physical articulation joint.

To handle the above challenges, we propose a framework towards \textbf{R}obust \textbf{P}erception and \textbf{M}anipulation for \textbf{Art}iculated Objects (\textbf{RPMArt}), which learns to estimate the articulation parameters and manipulate the articulation part from the noisy point cloud as depicted in Fig. \ref{fig:teaser}. We draw inspirations from BeyondPPF \cite{cppf, beyondppf}, a sim-to-real 9D object pose estimation method, which formulates the problem of pose estimation as a voting process. Given the point cloud, several point tuples are sampled, and a \textbf{Ro}bust \textbf{Art}iculation \textbf{Net}work (\textbf{RoArtNet}) is trained to generate the offsets to the articulation joints and affordable points from the local features of these point tuples. For each point tuple, RoArtNet votes several target candidates. After enumerating all the candidates, the target with the most votes is regarded as the final estimation. Moreover, we introduce an articulation-aware classification scheme to make RoArtNet aware of articulated objects' unique geometric structure, facilitating better sim-to-real transfer. Finally, AnyGrasp \cite{anygrasp} is used to propose a series of candidate grasp poses and the affordable initial grasp pose is selected based on the estimated affordable point. And the robot manipulation actions are guided by the estimated articulation joint using impedance control. RPMArt is trained only on synthetic data and is able to transfer zero-shot to real-world articulated objects. We conduct extensive experiments in both simulation and real-world environments, and achieve state-of-the-art performance.

Overall, our contributions are summarized as follows: 

\begin{itemize}
    \item We present RoArtNet, a robust articulation network that takes an articulation-aware voting approach based on local point tuple features to estimate joint parameters and affordable points robustly, facilitating effective transfer to real-world scenarios.
    \item We employ affordance-based, physics-guided manipulation to generate effective and robust actions executed by the robot, incorporating affordable grasp selection and articulation joint constraint.
    \item We conduct comprehensive experiments in both simulation and real world, and achieve state-of-the-art performance on both perception and manipulation tasks.
\end{itemize}

\section{Related Work}

\textbf{Articulation perception} has been studied for decades, where early stage methods often recover the poses for different parts with prior instance information available such as CAD models \cite{ocr, factored}. More recently, with the development of deep learning techniques, articulation perception from raw sensory data becomes possible. Hu et al. \cite{snapshot} introduced a part mobility model to map the single static snapshot to dynamic units in the training set. Though this querying method can achieve motion prediction and transfer to the input object, it needs part segmentation as prior information. Shape2Motion \cite{shape2motion} takes a two-stage method with mobility proposal and optimization networks to segment motion parts and estimate joint poses, but it is trained and tested on the whole point clouds. The following methods \cite{kno, ancsh, omad, gapartnet, gamma} exploit strong point cloud processing backbones \cite{mixture, pointnet++, sparseunet} to model articulated objects from single-view point clouds. Though they achieve accurate estimation on synthetic articulated objects, their generalization to the real-world cases is not guaranteed, especially in the presence of unexpected noise. This work deals with single-view real-world point clouds, with only synthetic data used for training.

\textbf{Articulated object manipulation} aims to manipulate the movable part of the articulated object by a robot, and prior works can be broadly categorized into learning-based and planning-based. Some learning-based methods leverage imitation learning \cite{clil, hiveformer} or reinforcement learning \cite{maniskill, maniskill2} to learn policy from collected robot demonstrations. However, collecting high-quality demonstrations is time-consuming and expensive. Another line of learning-based methods relies on learning visual affordance heatmap \cite{affordance-theory, affordance-survey} to select contact poses and predict actions \cite{where2act, vat-mart, umpnet}. However, the affordance heatmap is ambiguous and hard to annotate. On the other hand, planning-based methods often compute a motion trajectory with some geometry knowledge perfectly known \cite{door-open-control, door-open-plan} or estimated visually \cite{open-new-door, door-operation}. This work lies in the planning-based methods but also learns affordance to incorporate semantic understanding.

\textbf{Sim-to-real transfer} is commonly needed in many real-world application fields. Although there is vast literature on rigid objects pose estimation \cite{posecnn, nocs, cppf, beyondppf}, a few works have been devoted to articulated objects perception and manipulation. Like other fields, ReArtNOCS \cite{reartnocs} renders scanned articulated object models under different real scene backgrounds to synthesize data for training of articulation poses. However, it still does not take care of the domain gap between synthetic and real point clouds, and this tricky rendering process implicitly makes an assumption of test data distribution. This work draws inspirations from BeyondPPF \cite{beyondppf} and wants to narrow the sim-to-real gap for articulated objects perception and manipulation.

\section{Problem Formulation}

\begin{wrapfigure}{r}{0.48\columnwidth}
\centering
\includegraphics[width=0.48\columnwidth, trim={3mm 3mm 3mm 2mm},clip]{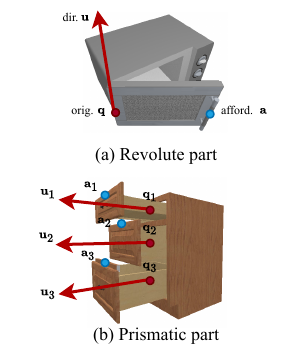}
\caption{\small Illustration of joint parameters and affordable points on articulated objects.}
\label{fig:problem}
\vspace{-2mm}
\end{wrapfigure}

Our perception goal is to estimate the joint parameters and affordable points from an observed articulated object's 3D point cloud $P \in \mathbb{R}^{N \times 3}$ backprojected from a single depth image, where $N$ denotes the number of points (see Fig. \ref{fig:problem}). Following other works \cite{ancsh, gamma}, we only consider 1D revolute joints and 1D prismatic joints. And we formulate the joint parameters as $\{\mathbf{u}_j, \mathbf{q}_j \mid j = 1,\ldots,J\}$, where $\mathbf{u}_j \in \mathbb{R}^3$ is the direction of the joint axis, $\mathbf{q}_j \in \mathbb{R}^3$ is the origin of the joint axis, and $J$ denotes the number of joints an articulated object comprises. Note that the origin of prismatic joint is also considered and defined as the center of part front surface in its rest state. Unlike previous works, affordable points $\{\mathbf{a}_j \in \mathbb{R}^3 \mid j = 1,\ldots,J\}$ instead of part segmentation or part bounding boxes are estimated. The affordable point represents the affordance \cite{affordance-theory, affordance-survey} peak among the space, indicating the potential interaction between robot and object which is the most likely to succeed. Our manipulation tasks include pulling and pushing the articulation part by a robot with a two-finger parallel gripper, while ensuring that the change of joint state exceeds a specific threshold.

\section{Method}

RPMArt uses RoArtNet, an articulation perception method to estimate joint parameters and affordable points from a noisy point cloud (depicted in Fig. \ref{fig:roartnet}), followed by an affordance-based, physics-guided manipulation pipeline (depicted in Fig. \ref{fig:teaser} (b)). First, several point tuples are sampled from the point cloud, and RoArtNet votes the joint parameters and affordable points based on these samples' local features (detailed in Sec. \ref{subsec:perception}). Moreover, RoArtNet is supervised by an articulation-aware classification loss during training, and selects votes with high articulation scores for sim-to-real transfer during inference (detailed in Sec. \ref{subsec:aware}). Finally, RPMArt selects an affordable initial grasp pose based on the estimated affordable point and executes subsequent actions constrained by the estimated joint parameters (detailed in Sec. \ref{subsec:manipulation}).

\subsection{RoArtNet for Point Tuple Voting}
\label{subsec:perception}

We draw inspirations from BeyondPPF \cite{beyondppf}, which is a sim-to-real rigid object pose estimation method that achieves state-of-the-art performance. Unlike most point cloud processing algorithms \cite{pointnet, pointnet++, pointmlp}, we want to refrain from aggregating global features of the whole point cloud and only rely on some distinctive local patterns. Thus, given the point cloud $P$, we sample $K$ point tuples from it, and each point tuple contains $M$ points, with the first two points $\mathbf{p}_1$ and $\mathbf{p}_2$ as the major points. For each point tuple $\mathcal{T} = \{\mathbf{p}_1, \ldots, \mathbf{p}_M\}$, we extract the following features as the network input: 
\begin{align}
& \mathcal{F}_1 = \operatorname{concat}(\{\mathbf{p}_i - \mathbf{p}_j \mid (i, j) \in \sigma^2(M)\}), \\
& \mathcal{F}_2 = \operatorname{concat}(\{\max{(\mathbf{n}_i \cdot \mathbf{n}_j, -\mathbf{n}_i \cdot \mathbf{n}_j)} \mid (i, j) \in \sigma^2(M)\}), \\
& \mathcal{F}_3 = \operatorname{concat}(\{\mathbf{s}'_i \mid i = 1, \ldots, M\}),
\end{align}
where $\operatorname{concat}$ means concatenation, $\sigma^2(M)$ represents all combinations of order $2$ from $M$ (a.k.a. $M$ choose $2$), $\{\mathbf{n}_1, \ldots, \mathbf{n}_M\}$ represents the normals of $\mathcal{T}$, and $\{\mathbf{s}'_1, \ldots, \mathbf{s}'_M\}$ is computed by MLP layers encoding the SHOT \cite{shot} features $\{\mathbf{s}_1, \ldots, \mathbf{s}_M\}$ of $\mathcal{T}$. Here, $\mathcal{F}_1$ represents the relative geometry information, while $\mathcal{F}_2$ and $\mathcal{F}_3$ contain the local context features around each point. Note that all these three features are translation invariant, while $\mathcal{F}_2$ and $\mathcal{F}_3$ are rotation invariant. Rendering under different camera poses can also make $\mathcal{F}_1$ rotation invariant. Such local features can help adapt to different situations, enhancing model robustness.

\begin{figure*}[!htbp]
\centering
\includegraphics[width=\textwidth]{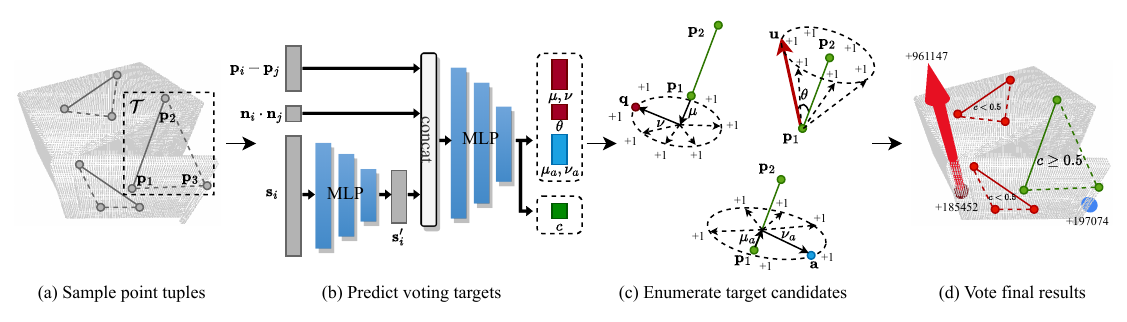}
\caption{\small Overview of RoArtNet. First, (a) a collection of $M$-point tuples ($M=3$ here as an example) are uniformly sampled from the point cloud. For each point tuple, (b) we predict several voting targets with a neural network from the local context features of the point tuple. Further, an articulation score $c$ is applied to supervise the neural network so that the network is aware of the articulation structure. Then, (c) we can generate multiple candidates using the predicted voting targets, given the one degree-of-freedom ambiguity constraint. (d) The candidate joint origin, joint direction and affordable point with the most votes, from only point tuples with high articulation score, are selected as the final estimation.}
\label{fig:roartnet}
\vspace{-5mm}
\end{figure*}

The network is implemented as a residual \cite{resnet} MLP, and predicts several offsets to the joint origin $\mathbf{q}$ and joint direction $\mathbf{u}$ with respect to the major points $\mathbf{p}_1$ and $\mathbf{p}_2$: 
\begin{align}
& \mu = \overrightarrow{\mathbf{p}_1 \mathbf{q}} \cdot \frac{\overrightarrow{\mathbf{p}_1 \mathbf{p}_2}}{\left\|\overrightarrow{\mathbf{p}_1 \mathbf{p}_2}\right\|_2}, \label{eq:mu} \\
& \nu = \left\|\mathbf{q}-\left(\mathbf{p}_1+\mu \frac{\overrightarrow{\mathbf{p}_1 \mathbf{p}_2}}{\left\|\overrightarrow{\mathbf{p}_1 \mathbf{p}_2}\right\|_2}\right)\right\|_2, \label{eq:nu} \\
& \theta = \mathbf{u} \cdot \frac{\overrightarrow{\mathbf{p}_1 \mathbf{p}_2}}{\left\|\overrightarrow{\mathbf{p}_1 \mathbf{p}_2}\right\|_2}. \label{eq:theta}
\end{align}
RoArtNet also predicts offsets to the affordable point $\mathbf{a}$ for subsequent grasp pose selection (see Sec. \ref{subsec:manipulation}): 
\begin{align}
& \mu_a = \overrightarrow{\mathbf{p}_1 \mathbf{a}} \cdot \frac{\overrightarrow{\mathbf{p}_1 \mathbf{p}_2}}{\left\|\overrightarrow{\mathbf{p}_1 \mathbf{p}_2}\right\|_2}, \\
& \nu_a = \left\|\mathbf{a}-\left(\mathbf{p}_1+\mu_a \frac{\overrightarrow{\mathbf{p}_1 \mathbf{p}_2}}{\left\|\overrightarrow{\mathbf{p}_1 \mathbf{p}_2}\right\|_2}\right)\right\|_2.
\end{align}
Once $\mu$ and $\nu$ are fixed, $\mathbf{q}$ is determined with up to one degree-of-freedom ambiguity in a circle, similar for $\mathbf{a}$ by $\mu_a$ and $\nu_a$. Similarly, once $\theta$ is fixed, $\mathbf{u}$ lies on a conical surface with one degree-of-freedom ambiguity. Thus, during inference, we can generate multiple candidates with a constant degree interval along the circle or cone for each point tuple, and the target will emerge with the most votes, as demonstrated in Fig. \ref{fig:roartnet} (c). Such voting scheme implicitly recognizes the distinctive local patterns, alleviating interference from noisy points.

For each point tuple $\mathcal{T}$, we optimize the joint origin loss $l^{\mathcal{T}}_{\text{orig}}$ and affordable point origin loss $l^{\mathcal{T}}_{\text{afford}}$ by mean squared error. Like other work \cite{classification}, we optimize the joint direction loss $l^{\mathcal{T}}_{\text{dir}}$ under the classification-based case by KL divergence. Formally, the vote loss for each tuple $\mathcal{T}$ is defined as: 
\begin{equation}
l^{\mathcal{T}}_{\text{vote}} = l^{\mathcal{T}}_{\text{orig}} + \lambda_d \cdot l^{\mathcal{T}}_{\text{dir}} + \lambda_a \cdot l^{\mathcal{T}}_{\text{afford}},
\end{equation}
where $\lambda_d$ and $\lambda_a$ are two weights to balance the influence of different terms.

\subsection{Articulation Awareness}
\label{subsec:aware}

After examining point clouds of both simulated and real-world articulated objects, we find that the distinctive articulation structure, featuring a movable part connected to the base either at an angle or with an offset, is shared commonly among various articulated objects and even preserves in the real-world noisy point clouds. Such a common structure can facilitate the generalization and sim-to-real transfer of the model. To make RoArtNet aware of the articulation structure, an additional articulation score $c_j$ is used to supervise the network during training. The ground truth articulation score $\{c_j \mid j = 1, \ldots, J\}$ of a sampled point tuple $\mathcal{T}$ is calculated based on the part segmentation $\{M_j \mid j = 0, \ldots, J\}$: 
\begin{equation}
c_j = \begin{cases}
1, &\text{if } (\mathbf{p}_1, \mathbf{p}_2) \in (M_0, M_j) \text{ or } (\mathbf{p}_1, \mathbf{p}_2) \in (M_j, M_0) \\
0, &\text{otherwise}
\end{cases},
\end{equation}
where $M_0$ denotes the mask of the base. This articulation score favors the point tuples whose two major points are located separately in the target part and the base. And the articulation awareness loss $\mathcal{L}_{\text{art}}$ is defined as the binary cross entropy between $c_j$ and predicted $\hat{c}_j$: 
\begin{equation}
\mathcal{L}_{\text{art}} = -\frac{1}{JK} \sum^J_{j=1} \sum_{\mathcal{T}_k} \left(c^k_j \log \hat{c}^k_j + (1 - c^k_j) \log (1 - \hat{c}^k_j)\right).
\end{equation}
During training, we only optimize the vote loss of the point tuples with articulation score as 1: 
\begin{equation}
\mathcal{L}_{\text{vote}} = \frac{1}{JC} \sum^J_{j=1} \sum_{\mathcal{T}_i} \{l^{\mathcal{T}_i}_{\text{vote}} \mid c^i_j = 1\},
\end{equation}
where $C = \operatorname{card}(\{l^{\mathcal{T}_i}_{\text{vote}} \mid c^i_j = 1\})$, and $\operatorname{card}(\cdot)$ means the cardinality of a set. Therefore, our final loss is defined as: 
\begin{equation}
\mathcal{L} = \mathcal{L}_{\text{vote}} + \lambda_{aa} \cdot \mathcal{L}_{\text{art}},
\end{equation}
where $\lambda_{aa}$ represents the weight of $\mathcal{L}_{\text{art}}$ term. And during inference, only the votes by point tuples with articulation score higher than 0.5 are kept for voting.

\subsection{Affordance-based Physics-guided Manipulation}
\label{subsec:manipulation}

To start manipulating the articulated object, the robot needs to first grasp the target part. We use AnyGrasp \cite{anygrasp} to generate a collection of grasp poses $\mathcal{G} = \{\mathbf{G}_g \in SE(3) \mid g = 1, \ldots, G\}$ given the point cloud $P$. In order to select one from these grasp poses that can manipulate the target part, we utilize the estimated affordable point. Currently, the grasp pose with minimum distance to the affordable point is selected. We define the affordable point for each part as the affordance \cite{affordance-theory, affordance-survey} peak among the part space, and we manually annotate the ground truth of affordable points. Typically, the affordable point lies on the edge center of the movable part, as shown in Fig. \ref{fig:problem}.

After grasping the target part, we explicitly exploit the estimated articulation joint and robot's proprioception to generate manipulation actions. In each time step $t$, we can sense current gripper pose $\mathbf{T}_t$ in robot base space and calculate the target pose $\hat{\mathbf{T}}_{t+1}$ with respect to the estimated articulation joint: 
\begin{equation}
\hat{\mathbf{T}}_{t+1} = \begin{cases}
    \operatorname{Rot}(\delta, \mathbf{u}, \mathbf{q}) \cdot \mathbf{T}_t,& \text{if revolute joint} \\
    \operatorname{Tr}(\delta, \mathbf{u}) \cdot \mathbf{T}_t,& \text{if prismatic joint}
\end{cases},
\end{equation}
where $\operatorname{Rot}(\delta, \mathbf{u}, \mathbf{q})$ represents the transformation matrix for rotating $\delta$ angle about axis $(\mathbf{u}, \mathbf{q})$, and $\operatorname{Tr}(\delta, \mathbf{u})$ represents the transformation matrix for translating $\delta$ distance along direction $\mathbf{u}$. We employ an impedance controller \cite{pid} to realize the actuation torques for reaching target poses.

\section{Experiments}

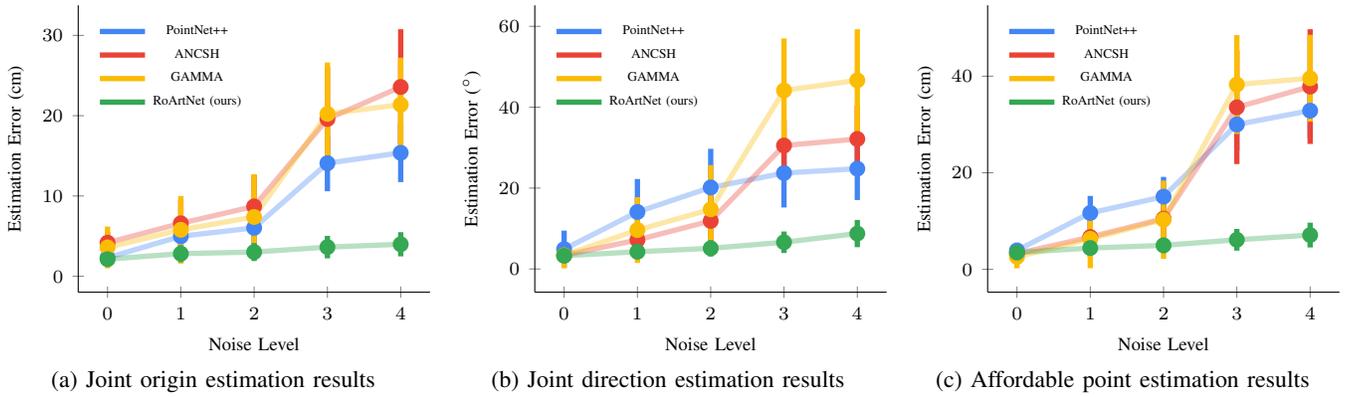
\begin{figure*}[!htbp]
\centering
\begin{subfigure}[c]{0.32\textwidth}
\centering
\begin{tikzpicture}
    \begin{axis}[
        width=1.1\textwidth,
        xlabel={\scriptsize Noise Level},
        ylabel={\scriptsize Estimation Error (cm)},
        legend pos=north west,
        grid=none,
        tick label style={font=\scriptsize},
        legend style={font=\tiny, draw=none, fill=none},
        xlabel style={label distance=0pt},
        ylabel style={label distance=0pt},
        axis x line=bottom,
        axis y line=left,
        axis line style={-},
        xtick=data,
        enlargelimits=auto,
        error bars/error bar style={line width=2pt},
        ]

        \addplot+[forget plot,mark=none,line width=2pt,draw opacity=0.35,color=myblue] coordinates {
            (0, 2.311156176)
            (1, 4.982594756)
            (2, 6.032397794)
            (3, 14.08432586)
            (4, 15.37625444)
        };
        \addplot+[forget plot,mark=*,line width=2pt,only marks,color=myblue,mark options={color=myblue},error bars/.cd,y dir=both,y explicit,error mark=none] coordinates {
            (0, 2.311156176) +- (1.668350631/2, 1.668350631/2)
            (1, 4.982594756) +- (3.978339172/2, 3.978339172/2)
            (2, 6.032397794) +- (4.549457764/2, 4.549457764/2)
            (3, 14.08432586) +- (6.991432103/2, 6.991432103/2)
            (4, 15.37625444) +- (7.299133936/2, 7.299133936/2)
        };
        \addlegendimage{mark=none,line width=2pt,color=myblue}
        \addlegendentry{PointNet++}

        \addplot+[forget plot,mark=none,line width=2pt,draw opacity=0.35,color=myred] coordinates {
            (0, 4.184292475)
            (1, 6.597484611)
            (2, 8.701421981)
            (3, 19.59020633)
            (4, 23.57343592)
        };
        \addplot+[forget plot,mark=*,line width=2pt,only marks,color=myred,mark options={color=myred},error bars/.cd,y dir=both,y explicit,error mark=none] coordinates {
            (0, 4.184292475) +- (2.67506157/2, 2.67506157/2)
            (1, 6.597484611) +- (5.865363861/2, 5.865363861/2)
            (2, 8.701421981) +- (7.822206828/2, 7.822206828/2)
            (3, 19.59020633) +- (12.94553611/2, 12.94553611/2)
            (4, 23.57343592) +- (14.38822458/2, 14.38822458/2)
        };
        \addlegendimage{mark=none,line width=2pt,color=myred}
        \addlegendentry{ANCSH}

        \addplot+[forget plot,mark=none,line width=2pt,draw opacity=0.35,color=myyellow] coordinates {
            (0, 3.594448082)
            (1, 5.788608881)
            (2, 7.385695761)
            (3, 20.22442275)
            (4, 21.37634104)
        };
        \addplot+[forget plot,mark=*,line width=2pt,only marks,color=myyellow,mark options={color=myyellow},error bars/.cd,y dir=both,y explicit,error mark=none] coordinates {
            (0, 3.594448082) +- (5.194149292/2, 5.194149292/2)
            (1, 5.788608881) +- (8.416512728/2, 8.416512728/2)
            (2, 7.385695761) +- (10.65655019/2, 10.65655019/2)
            (3, 20.22442275) +- (12.79322253/2, 12.79322253/2)
            (4, 21.37634104) +- (11.69984658/2, 11.69984658/2)
        };
        \addlegendimage{mark=none,line width=2pt,color=myyellow}
        \addlegendentry{GAMMA}

        \addplot+[forget plot,mark=none,line width=2pt,draw opacity=0.35,color=mygreen] coordinates {
            (0, 2.129139417)
            (1, 2.817760944)
            (2, 3.019536097)
            (3, 3.635606558)
            (4, 3.987097539)
        };
        \addplot+[forget plot,mark=*,line width=2pt,only marks,color=mygreen,mark options={color=mygreen},error bars/.cd,y dir=both,y explicit,error mark=none] coordinates {
            (0, 2.129139417) +- (1.59744437/2, 1.59744437/2)
            (1, 2.817760944) +- (2.122744097/2, 2.122744097/2)
            (2, 3.019536097) +- (2.238422764/2, 2.238422764/2)
            (3, 3.635606558) +- (2.806904786/2, 2.806904786/2)
            (4, 3.987097539) +- (3.014992994/2, 3.014992994/2)
        };
        \addlegendimage{mark=none,line width=2pt,color=mygreen}
        \addlegendentry{RoArtNet (ours)}
        
    \end{axis}
\end{tikzpicture}
\vspace{-5mm}
\caption{Joint origin estimation results}
\label{fig:sim-orig}
\end{subfigure}
\hfill
\begin{subfigure}[c]{0.32\textwidth}
\centering
\begin{tikzpicture}
    \begin{axis}[
        width=1.1\textwidth,
        xlabel={\scriptsize Noise Level},
        ylabel={\scriptsize Estimation Error ($^\circ$)},
        legend pos=north west,
        grid=none,
        tick label style={font=\scriptsize},
        legend style={font=\tiny, draw=none, fill=none},
        xlabel style={label distance=0pt},
        ylabel style={label distance=0pt},
        axis x line=bottom,
        axis y line=left,
        axis line style={-},
        xtick=data,
        enlargelimits=auto,
        error bars/error bar style={line width=2pt},
        ]

        \addplot+[forget plot,mark=none,line width=2pt,draw opacity=0.35,color=myblue] coordinates {
            (0, 4.897692533)
            (1, 14.09573569)
            (2, 20.15249492)
            (3, 23.7130075)
            (4, 24.79083947)
        };
        \addplot+[forget plot,mark=*,line width=2pt,only marks,color=myblue,mark options={color=myblue},error bars/.cd,y dir=both,y explicit,error mark=none] coordinates {
            (0, 4.897692533) +- (9.180635347/2, 9.180635347/2)
            (1, 14.09573569) +- (16.30497073/2, 16.30497073/2)
            (2, 20.15249492) +- (19.11599606/2, 19.11599606/2)
            (3, 23.7130075) +- (17.01891362/2, 17.01891362/2)
            (4, 24.79083947) +- (15.51382633/2, 15.51382633/2)
        };
        \addlegendimage{mark=none,line width=2pt,color=myblue}
        \addlegendentry{PointNet++}

        \addplot+[forget plot,mark=none,line width=2pt,draw opacity=0.35,color=myred] coordinates {
            (0, 3.367602964)
            (1, 7.177231886)
            (2, 11.88606428)
            (3, 30.51755597)
            (4, 32.14240897)
        };
        \addplot+[forget plot,mark=*,line width=2pt,only marks,color=myred,mark options={color=myred},error bars/.cd,y dir=both,y explicit,error mark=none] coordinates {
            (0, 3.367602964) +- (3.039114372/2, 3.039114372/2)
            (1, 7.177231886) +- (7.898357314/2, 7.898357314/2)
            (2, 11.88606428) +- (12.64478101/2, 12.64478101/2)
            (3, 30.51755597) +- (12.60948725/2, 12.60948725/2)
            (4, 32.14240897) +- (16.5373795/2, 16.5373795/2)
        };
        \addlegendimage{mark=none,line width=2pt,color=myred}
        \addlegendentry{ANCSH}

        \addplot+[forget plot,mark=none,line width=2pt,draw opacity=0.35,color=myyellow] coordinates {
            (0, 3.387817375)
            (1, 9.619453831)
            (2, 14.7364155)
            (3, 44.14338169)
            (4, 46.64980033)
        };
        \addplot+[forget plot,mark=*,line width=2pt,only marks,color=myyellow,mark options={color=myyellow},error bars/.cd,y dir=both,y explicit,error mark=none] coordinates {
            (0, 3.387817375) +- (6.436276339/2, 6.436276339/2)
            (1, 9.619453831) +- (16.25409611/2, 16.25409611/2)
            (2, 14.7364155) +- (21.91795856/2, 21.91795856/2)
            (3, 44.14338169) +- (25.73623622/2, 25.73623622/2)
            (4, 46.64980033) +- (25.26099617/2, 25.26099617/2)
        };
        \addlegendimage{mark=none,line width=2pt,color=myyellow}
        \addlegendentry{GAMMA}

        \addplot+[forget plot,mark=none,line width=2pt,draw opacity=0.35,color=mygreen] coordinates {
            (0, 3.271374194)
            (1, 4.253546289)
            (2, 5.106569156)
            (3, 6.609038619)
            (4, 8.754608469)
        };
        \addplot+[forget plot,mark=*,line width=2pt,only marks,color=mygreen,mark options={color=mygreen},error bars/.cd,y dir=both,y explicit,error mark=none] coordinates {
            (0, 3.271374194) +- (2.494281672/2, 2.494281672/2)
            (1, 4.253546289) +- (3.337747964/2, 3.337747964/2)
            (2, 5.106569156) +- (4.031089031/2, 4.031089031/2)
            (3, 6.609038619) +- (5.315807022/2, 5.315807022/2)
            (4, 8.754608469) +- (6.741201519/2, 6.741201519/2)
        };
        \addlegendimage{mark=none,line width=2pt,color=mygreen}
        \addlegendentry{RoArtNet (ours)}
        
    \end{axis}
\end{tikzpicture}
\vspace{-5mm}
\caption{Joint direction estimation results}
\label{fig:sim-dir}
\end{subfigure}
\hfill
\begin{subfigure}[c]{0.32\textwidth}
\centering
\begin{tikzpicture}
    \begin{axis}[
        width=1.1\textwidth,
        xlabel={\scriptsize Noise Level},
        ylabel={\scriptsize Estimation Error (cm)},
        legend pos=north west,
        grid=none,
        tick label style={font=\scriptsize},
        legend style={font=\tiny, draw=none, fill=none},
        xlabel style={label distance=0pt},
        ylabel style={label distance=0pt},
        axis x line=bottom,
        axis y line=left,
        axis line style={-},
        xtick=data,
        enlargelimits=auto,
        error bars/error bar style={line width=2pt},
        ]

        \addplot+[forget plot,mark=none,line width=2pt,draw opacity=0.35,color=myblue] coordinates {
            (0, 3.928740317)
            (1, 11.67293875)
            (2, 15.04824611)
            (3, 29.99894428)
            (4, 32.85597722)
        };
        \addplot+[forget plot,mark=*,line width=2pt,only marks,color=myblue,mark options={color=myblue},error bars/.cd,y dir=both,y explicit,error mark=none] coordinates {
            (0, 3.928740317) +- (2.112226203/2, 2.112226203/2)
            (1, 11.67293875) +- (7.071774133/2, 7.071774133/2)
            (2, 15.04824611) +- (8.205774944/2, 8.205774944/2)
            (3, 29.99894428) +- (10.83473232/2, 10.83473232/2)
            (4, 32.85597722) +- (11.32797751/2, 11.32797751/2)
        };
        \addlegendimage{mark=none,line width=2pt,color=myblue}
        \addlegendentry{PointNet++}

        \addplot+[forget plot,mark=none,line width=2pt,draw opacity=0.35,color=myred] coordinates {
            (0, 3.244880261)
            (1, 6.691246586)
            (2, 10.50375211)
            (3, 33.55373006)
            (4, 37.84380639)
        };
        \addplot+[forget plot,mark=*,line width=2pt,only marks,color=myred,mark options={color=myred},error bars/.cd,y dir=both,y explicit,error mark=none] coordinates {
            (0, 3.244880261) +- (1.813516503/2, 1.813516503/2)
            (1, 6.691246586) +- (8.132254678/2, 8.132254678/2)
            (2, 10.50375211) +- (12.48968119/2, 12.48968119/2)
            (3, 33.55373006) +- (23.50449669/2, 23.50449669/2)
            (4, 37.84380639) +- (23.75981231/2, 23.75981231/2)
        };
        \addlegendimage{mark=none,line width=2pt,color=myred}
        \addlegendentry{ANCSH}

        \addplot+[forget plot,mark=none,line width=2pt,draw opacity=0.35,color=myyellow] coordinates {
            (0, 2.583951035)
            (1, 6.316600858)
            (2, 10.28328711)
            (3, 38.27178803)
            (4, 39.56611831)
        };
        \addplot+[forget plot,mark=*,line width=2pt,only marks,color=myyellow,mark options={color=myyellow},error bars/.cd,y dir=both,y explicit,error mark=none] coordinates {
            (0, 2.583951035) +- (4.721275367/2, 4.721275367/2)
            (1, 6.316600858) +- (12.15343606/2, 12.15343606/2)
            (2, 10.28328711) +- (16.20190986/2, 16.20190986/2)
            (3, 38.27178803) +- (20.48964786/2, 20.48964786/2)
            (4, 39.56611831) +- (17.9984785/2, 17.9984785/2)
        };
        \addlegendimage{mark=none,line width=2pt,color=myyellow}
        \addlegendentry{GAMMA}

        \addplot+[forget plot,mark=none,line width=2pt,draw opacity=0.35,color=mygreen] coordinates {
            (0, 3.480443186)
            (1, 4.416817731)
            (2, 4.969218294)
            (3, 6.122470108)
            (4, 7.099475867)
        };
        \addplot+[forget plot,mark=*,line width=2pt,only marks,color=mygreen,mark options={color=mygreen},error bars/.cd,y dir=both,y explicit,error mark=none] coordinates {
            (0, 3.480443186) +- (1.92528875/2, 1.92528875/2)
            (1, 4.416817731) +- (2.758234778/2, 2.758234778/2)
            (2, 4.969218294) +- (3.157588569/2, 3.157588569/2)
            (3, 6.122470108) +- (4.495115697/2, 4.495115697/2)
            (4, 7.099475867) +- (5.133482167/2, 5.133482167/2)
        };
        \addlegendimage{mark=none,line width=2pt,color=mygreen}
        \addlegendentry{RoArtNet (ours)}
        
    \end{axis}
\end{tikzpicture}
\vspace{-5mm}
\caption{Affordable point estimation results}
\label{fig:sim-afford}
\end{subfigure}
\caption{\small Articulation perception results. We gradually add higher level of noise to the input point clouds, and test the joint parameters and affordable points estimation performance. Lower is better. Results are averaged across six object categories. Error bars represent the standard deviation. Different noise levels are detailed in Sec. \ref{parag:noise-perception}. More detailed results for each category are listed on our website.}
\label{fig:sim-perception}
\vspace{-5mm}
\end{figure*}

We perform our experiments in both simulated and real-world environments, and validate our framework by answering the following questions: (i) Can RoArtNet robustly estimate joint parameters and affordable points from point clouds with different levels of noise? (ii) Can RPMArt still manipulate articulated objects successfully under observation noise? (iii) Can RPMArt transfer zero-shot to real-world articulated objects?

\subsection{Environmental Setup}

{\bf Settings.}
We conduct the simulated experiments in the SAPIEN simulator \cite{sapien}, which supports physical simulation for robots and articulated objects interaction. It also provides depth map and part-level information rendering. We use a Panda flying two-finger parallel gripper to perform the manipulation tasks. In our real-world environment, a 7-DOF Franka Emika robot arm with an Intel RealSense L515 LiDAR camera mounted on the robot’s wrist is used to observe and manipulate real-world articulated objects. Computing is done on a NVIDIA A100 GPU.

{\bf Datasets.}
In total, we use 74 synthetic objects in 6 selected categories from PartNet-Mobility \cite{shapenet, partnet}. And we randomly split them into training and testing instances. For each instance, we import it into SAPIEN simulator, scale it into normal object size with [0.8, 1.1] additional varying range, and randomly set joint states within the joint limit ranges. A camera with $640 \times 480$ resolution is used to capture the depth map and part-level mask. We spherically sample the camera viewpoint in front of the target object, with camera looking at the center of the target object. For the spherical sampling, we set the range of the azimuth angle to [-60$^\circ$, 60$^\circ$], and the elevation angle to [0$^\circ$, 60$^\circ$]. The distance between camera and object is uniformly distributed in [0.6, 1.2] meter. In practice, we sample 40 different states for each object and 5 camera views for each state to render data. Additionally, we also collect one real object instance for each selected category, and capture its point cloud under different joint states and camera views. Note that we only use the synthetic training instances for training, and conduct evaluation over the synthetic testing instances and real-world objects.

{\bf Implementation details.}
We set the number of sampled point tuples $K$ to 100,000 and each point tuple contains $M=5$ points. And we set the loss weights $\lambda_d=0.1$, $\lambda_a=1.0$ and $\lambda_{aa}=0.5$ in all our implementation.

{\bf Baselines.}
We compare our method to three baselines: (i) a naive PointNet++ \cite{pointnet++} that takes the point cloud as input and directly outputs the joint parameters and affordable points; (ii) ANCSH \cite{ancsh} that exploits normalized coordinate space to estimate joint parameters and uses RANSAC \cite{ransac} to optimize the transformation to the camera space; (iii) GAMMA \cite{gamma} that learns dense projection offsets to vote joint parameters and dense clustering offsets to group part points. For ANCSH and GAMMA, we mimic their joint origin estimation and add an additional head for affordable point estimation. And we also use their perception results to finish manipulation tasks.

\begin{figure*}[!htbp]
\centering
\def\axisdefaultwidth{\textwidth}
\def\axisdefaultheight{160pt}
\begin{tikzpicture}
\pgfplotsset{
    every axis/.style={
        ybar stacked,
        ymin=0,ymax=100,
        ylabel={Success Rate (\%)},
        xtick={1,2,3,4,5,6},
        xticklabels={\footnotesize Pull microwave, \footnotesize Push microwave, \footnotesize Pull prismatic storage, \footnotesize Push prismatic storage, \footnotesize Pull revolute storage, \footnotesize Push revolute storage},
        ymajorgrids=true,
        axis x line=bottom,
        axis y line=left,
        axis line style={-},
        enlarge x limits=0.1,
        bar width=12pt,
    },
}

\begin{axis}[bar shift=-24pt,legend style={draw=none, fill=none, at={(0.2,-0.2)}, anchor=north}]
\addplot[draw=myblue,fill=myblue] coordinates
{(1,0.187165775*100) (2,0.728682171*100) (3,0.376344086*100) (4,0.621212121*100) (5,0.269430052*100) (6,0.806451613*100)};
\addlegendentry{PointNet++}
\addplot[draw=myblue!0,fill=myblue!67,pattern=crosshatch,pattern color=myblue!67, forget plot] coordinates
{(1,0.481405653*100) (2,0.099676039*100) (3,0.075432564*100) (4,-0.113965744*100) (5,0.230569948*100) (6,0.02688172*100)};
\addplot[draw=myblue!0,fill=myblue!33,pattern=north west lines,pattern color=myblue!33, forget plot] coordinates
{(1,0.205165945*100) (2,0.132579291*100) (3,0.40148422*100) (4,0.377999525*100) (5,0.458333333*100) (6,0.166666667*100)};
\end{axis}

\begin{axis}[bar shift=-8pt,hide axis,legend style={draw=none, fill=none, at={(0.4,-0.2)}, anchor=north}]
\addplot+[draw=myred,fill=myred] coordinates
{(1,0.448275862*100) (2,0.496503497*100) (3,0.005208333*100) (4,0*100) (5,0.436170213*100) (6,0.581081081*100)};
\addlegendentry{ANCSH}
\addplot+[draw=myred!0,fill=myred!67,pattern=crosshatch,pattern color=myred!67, forget plot] coordinates
{(1,0.149425287*100) (2,0.261885765*100) (3,0.596942204*100) (4,0.62745098*100) (5,0.115376179*100) (6,0.365977742*100)};
\addplot+[draw=myred!0,fill=myred!33,pattern=north west lines,pattern color=myred!33, forget plot] coordinates
{(1,0.105423851*100) (2,-0.000813504*100) (3,0.347344412*100) (4,0.324929972*100) (5,0.066254655*100) (6,0.033710407*100)};
\end{axis}

\begin{axis}[bar shift=8pt,hide axis,legend style={draw=none, fill=none, at={(0.6,-0.2)}, anchor=north}]
\addplot+[draw=myyellow,fill=myyellow] coordinates
{(1,0.415300546*100) (2,0.620155039*100) (3,0.231182796*100) (4,0.318181818*100) (5,0.419689119*100) (6,0.64516129*100)};
\addlegendentry{GAMMA}
\addplot+[draw=myyellow!0,fill=myyellow!67,pattern=crosshatch,pattern color=myyellow!67, forget plot] coordinates
{(1,0.321842311*100) (2,0.256656555*100) (3,0.677446646*100) (4,0.638339921*100) (5,0.460118573*100) (6,0.336320191*100)};
\addplot+[draw=myyellow!0,fill=myyellow!33,pattern=north west lines,pattern color=myyellow!33, forget plot] coordinates
{(1,0.171948052*100) (2,0.091938406*100) (3,0.03158795*100) (4,0.010691376*100) (5,0.099358974*100) (6,0.012066906*100)};
\end{axis}

\begin{axis}[bar shift=24pt,hide axis,legend style={draw=none, fill=none, at={(0.8,-0.2)}, anchor=north}]
\addplot+[draw=mygreen,fill=mygreen] coordinates
{(1,0.76331361*100) (2,0.942028986*100) (3,0.827225131*100) (4,1*100) (5,0.823834197*100) (6,0.981012658*100)};
\addlegendentry{RPMArt (ours)}
\addplot+[draw=mygreen!0,fill=mygreen!67,pattern=crosshatch,pattern color=mygreen!67, forget plot] coordinates
{(1,0.126221274*100) (2,0.036542443*100) (3,0.117498487*100) (4,0*100) (5,0.03712837*100) (6,0.005654008*100)};
\addplot+[draw=mygreen!0,fill=mygreen!33,pattern=north west lines,pattern color=mygreen!33, forget plot] coordinates
{(1,-0.001105132*100) (2,-0.021736896*100) (3,0.013609715*100) (4,-0.017241379*100) (5,0.061689367*100) (6,0.000675105*100)};
\end{axis}

\begin{scope}[shift={(current bounding box.south)}, yshift=17.5pt, xshift=-100pt]
  \draw[draw=gray!0, pattern=north west lines, pattern color=gray!33] (0,0) rectangle (0.25,0.3) node[right] at (0.2,0.15) {\scriptsize Noise Level 0};
\end{scope}
\begin{scope}[shift={(current bounding box.south)}, yshift=17.5pt]
  \draw[draw=gray!0, pattern=crosshatch, pattern color=gray!67] (0,0) rectangle (0.25,0.3) node[right] at (0.2,0.15) {\scriptsize Noise Level 2};
\end{scope}
\begin{scope}[shift={(current bounding box.south)}, yshift=17.5pt, xshift=100pt]
  \draw[draw=gray, fill=gray] (0,0) rectangle (0.25,0.3) node[right] at (0.2,0.15) {\scriptsize Noise Level 4};
\end{scope}

\end{tikzpicture}
\vspace{-5mm}
\caption{\small Articulated object manipulation results. We report the success rate averaged among around 100 trials per object instance for each task. Higher is better. Selected noise levels are detailed in Sec. \ref{parag:noise-manipulation}. More results for other tasks are shown on our website.}
\label{fig:sim-manipulation}
\vspace{-5mm}
\end{figure*}
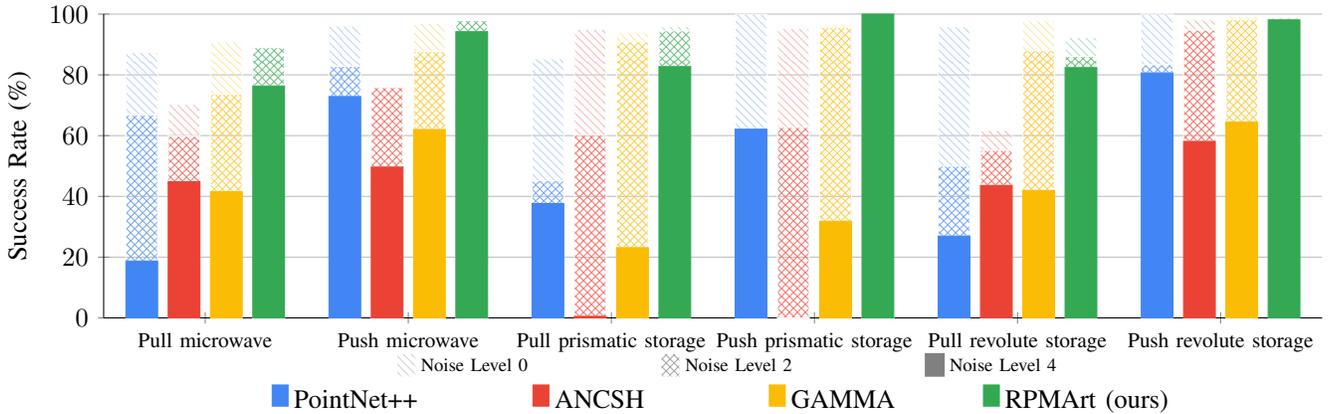

\subsection{Articulation Perception Results}

{\bf Metrics.}
We evaluate the orientation error of the joint axis direction in degrees. We evaluate the translation error of the joint axis origin using the minimum line-to-line distance in centimeters for revolute joints, and using L2 distance in centimeters for prismatic joints. And we evaluate the translation error of the affordable point using L2 distance in centimeters.

{\bf Results.}
\label{parag:noise-perception}
To validate the robustness of our method, we test the models on the point clouds with different levels of noise. Like PointCleanNet \cite{pointcleannet}, we add two types of noise to raw point clouds. For the distortion noise, a certain percentage $\rho_d$ of points are sampled and added Gaussian noise with the standard deviation as the proportion $\sigma_d$ of the original point cloud’s bounding box diagonal. And for the outlier noise, a certain percentage $\rho_o$ of points are sampled and replaced with random points that are generated by uniformly sampling among a larger bounding box, whose size is the proportion $\sigma_o$ of the original bounding box, with the same center. In our experiments, five levels of $(\rho_d, \sigma_d, \rho_o, \sigma_o)$ are tested, as (0, 0, 0, 0), (0.1, 0.01, 0.001, 1.0), (0.2, 0.01, 0.002, 1.0), (0.1, 0.02, 0.001, 2.0), and (0.2, 0.02, 0.002, 2.0). Fig. \ref{fig:sim-perception} presents the articulation perception results. Results show that all baselines and RoArtNet achieve high estimation precision without noise added. Nevertheless, with the increasing level of noise, all three baselines exhibit a pronounced increase in estimation errors, with $5.1\times$ to $15.3\times$ increase. In contrast, the mean estimation error of RoArtNet increases very slowly, with $1.9\times$ to $2.7\times$ increase. And the baselines also have much higher standard deviation compared to RoArtNet when high level of noise is added. In addition, we also conduct an ablation study to analyze the influence of articulation awareness. We use votes from all point tuples to determine the final estimation, rather than discarding point tuples with low articulation score. The performance within the Microwave category under noise level 2 is shown in Table \ref{tab:ablation-perception}. The performance degrades, especially for affordable point estimation, underscoring the robustness brought by articulation awareness.

\begin{table}[!htbp]
\centering
\caption{\small Ablation study on articulation awareness.}
\label{tab:ablation-perception}
\begin{tabular}{cccc}
\toprule
\multirow{2}{*}{Model} & \multicolumn{3}{c}{Error $\downarrow$} \\
 & Orig. (cm) & Dir. ($^\circ$) & Afford. (cm) \\
\midrule
Ours w/o awareness & 4.81{\tiny $\pm$3.97} & 5.04{\tiny $\pm$4.03} & 13.33{\tiny $\pm$10.10} \\
Ours (full) & \textbf{3.34}{\tiny $\pm$\textbf{3.11}} & \textbf{4.53}{\tiny $\pm$\textbf{3.84}} & \textbf{6.62}{\tiny $\pm$\textbf{4.88}} \\
\bottomrule
\end{tabular}

\vspace{-3mm}   %
\end{table}

\subsection{Articulated Object Manipulation Results}

{\bf Metrics.}
We run around 100 interaction trials per articulated object instance and report success rates of changing the target joint state over a threshold ratio (here we set 0.85) of specific task value (here we randomly choose a rate from [0.1, 0.7] of the joint limit).

{\bf Results.}
\label{parag:noise-manipulation}
Like in the articulation perception experiments, we also add noise to the observed point clouds with different levels of $(\rho_d, \sigma_d, \rho_o, \sigma_o)$, as (0, 0, 0, 0), (0.2, 0.01, 0.002, 1.0), and (0.2, 0.02, 0.002, 2.0). Fig. \ref{fig:sim-manipulation} shows six example task results. It is clear that our method achieves the highest success rate under noise level 4 across all tasks. And we can observe the least degradation in performance of our method with the increase of noise. In addition, we also implement ablation studies to validate different components of our manipulation within the Microwave category under noise level 2, as presented in Table \ref{tab:ablation-manipulation}. We first use the grasp score predicted by AnyGrasp instead of the estimated affordable point to select the initial grasp pose. The performance degrades to much lower success rates, highlighting the importance of affordance-based semantic understanding. Then, we also attempt to plan the entire trajectory at the initial stage instead of constraining by the estimated joint parameters in each time step. The success rate also decreases, indicating the necessity of physical constraints by the articulation joint.

\begin{table}[!htbp]
\centering
\caption{\small Ablation studies on our affordance-based physics-guided manipulation.}
\label{tab:ablation-manipulation}
\begin{tabular}{cccc}
\toprule
\multirow{2}{*}{Method} & \multicolumn{2}{c}{Success rate (\%) $\uparrow$} \\
 & Pull & Push \\
\midrule
Ours w/o affordance & 38.953 & 29.286 \\
Ours w/o constraint & 77.326 & 95.714 \\
Ours (full) & \textbf{88.953} & \textbf{97.857} \\
\bottomrule
\end{tabular}

\vspace{-2mm}
\end{table}

\subsection{Real-world Experiments}

\begin{table}[!htbp]
\centering
\caption{\small Quantitative evaluation of the performance on real-world articulation perception.}
\label{tab:real-perception}

\setlength{\tabcolsep}{4pt}
\begin{tabular}{ccccc}
\toprule
\multirow{2}{*}{Category} & \multirow{2}{*}{Method} & \multicolumn{3}{c}{Error $\downarrow$} \\
 & & Orig. (cm) & Dir. ($^\circ$) & Afford. (cm) \\
\midrule
\multirow{4}{*}{Microwave} & PointNet++ \cite{pointnet++} & 4.49{\tiny $\pm$3.57} & 9.27{\tiny $\pm$5.83} & 15.44{\tiny $\pm$4.73} \\
 & ANCSH \cite{ancsh} & 5.10{\tiny $\pm$5.52} & 9.17{\tiny $\pm$9.56} & 12.71{\tiny $\pm$7.93} \\
 & GAMMA \cite{gamma} & \textbf{2.53}{\tiny $\pm$2.90} & 9.91{\tiny $\pm$10.67} & 7.24{\tiny $\pm$10.19} \\
 & RoArtNet (ours) & 3.83{\tiny $\pm$\textbf{2.37}} & \textbf{5.19}{\tiny $\pm$\textbf{3.62}} & \textbf{6.75}{\tiny $\pm$\textbf{3.28}} \\
\hline
\multirow{4}{*}{Refrigerator} & PointNet++ \cite{pointnet++} & 5.21{\tiny $\pm$4.27} & 9.60{\tiny $\pm$5.34} & 12.47{\tiny $\pm$9.50} \\
 & ANCSH \cite{ancsh} & 5.94{\tiny $\pm$5.80} & \textbf{8.00}{\tiny $\pm$5.91} & 12.81{\tiny $\pm$13.60} \\
 & GAMMA \cite{gamma} & 4.02{\tiny $\pm$4.58} & 8.68{\tiny $\pm$6.46} & 12.33{\tiny $\pm$9.97} \\
 & RoArtNet (ours) & \textbf{2.11}{\tiny $\pm$\textbf{1.70}} & 8.49{\tiny $\pm$\textbf{4.27}} & \textbf{5.85}{\tiny $\pm$\textbf{2.80}} \\
\hline
\multirow{4}{*}{Safe} & PointNet++ \cite{pointnet++} & 5.99{\tiny $\pm$4.16} & 5.94{\tiny $\pm$2.86} & 9.23{\tiny $\pm$5.63} \\
 & ANCSH \cite{ancsh} & 5.17{\tiny $\pm$6.76} & 7.71{\tiny $\pm$14.28} & 8.51{\tiny $\pm$9.77} \\
 & GAMMA \cite{gamma} & \textbf{3.18}{\tiny $\pm$3.86} & 8.16{\tiny $\pm$13.74} & 9.06{\tiny $\pm$9.67} \\
 & RoArtNet (ours) & 4.12{\tiny $\pm$\textbf{2.43}} & \textbf{5.88}{\tiny $\pm$\textbf{2.77}} & \textbf{8.35}{\tiny $\pm$\textbf{4.39}} \\
\hline
\multirow{4}{*}{\makecell{Storage\\Furniture}} & PointNet++ \cite{pointnet++} & 7.54{\tiny $\pm$4.52} & \textbf{8.78}{\tiny $\pm$\textbf{4.99}} & 10.63{\tiny $\pm$4.03} \\
 & ANCSH \cite{ancsh} & 6.41{\tiny $\pm$4.22} & 9.61{\tiny $\pm$6.40} & 5.18{\tiny $\pm$6.02} \\
 & GAMMA \cite{gamma} & \textbf{3.48}{\tiny $\pm$2.28} & 12.67{\tiny $\pm$10.19} & \textbf{4.74}{\tiny $\pm$6.66} \\
 & RoArtNet (ours) & 4.60{\tiny $\pm$\textbf{2.05}} & 9.68{\tiny $\pm$5.45} & 7.945{\tiny $\pm$\textbf{3.40}} \\
\hline
\multirow{4}{*}{Drawer} & PointNet++ \cite{pointnet++} & 8.33{\tiny $\pm$3.38} & \textbf{7.86}{\tiny $\pm$\textbf{5.30}} & 10.23{\tiny $\pm$4.46} \\
 & ANCSH \cite{ancsh} & 13.85{\tiny $\pm$3.76} & 12.14{\tiny $\pm$8.03} & 7.72{\tiny $\pm$4.70} \\
 & GAMMA \cite{gamma} & \textbf{5.06}{\tiny $\pm$\textbf{2.36}} & 14.67{\tiny $\pm$6.77} & \textbf{6.97}{\tiny $\pm$\textbf{3.11}} \\
 & RoArtNet (ours) & 5.99{\tiny $\pm$3.06} & 11.31{\tiny $\pm$5.60} & 7.73{\tiny $\pm$5.25} \\
\hline
\multirow{4}{*}{\makecell{Washing\\Machine}} & PointNet++ \cite{pointnet++} & 8.85{\tiny $\pm$6.80} & 37.50{\tiny $\pm$20.68} & 19.97{\tiny $\pm$9.44} \\
 & ANCSH \cite{ancsh} & 5.16{\tiny $\pm$4.92} & 16.24{\tiny $\pm$12.09} & 11.54{\tiny $\pm$8.23} \\
 & GAMMA \cite{gamma} & 6.49{\tiny $\pm$6.18} & 28.44{\tiny $\pm$14.87} & 15.96{\tiny $\pm$13.29} \\
 & RoArtNet (ours) & \textbf{1.58}{\tiny $\pm$\textbf{1.20}} & \textbf{5.60}{\tiny $\pm$\textbf{2.71}} & \textbf{3.25}{\tiny $\pm$\textbf{0.67}} \\
\bottomrule
\end{tabular}

\end{table}

\begin{figure}[!htbp]
\centering
\includegraphics[width=\columnwidth]{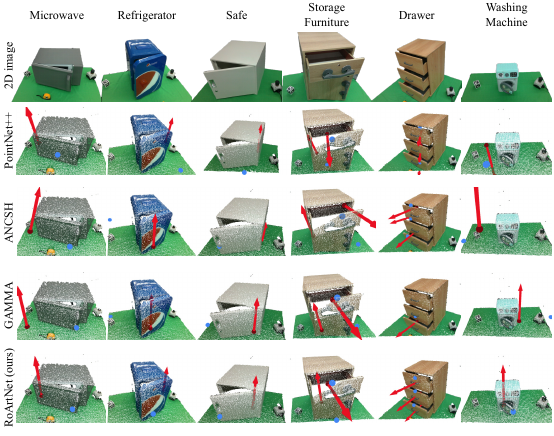}
\caption{\small Qualitative results of the performance on real-world articulation perception. Color is used only for visualization here. Red arrows represent the estimated articulation joints, and blue points represent the estimated affordable points. Zoom in for better view.}
\label{fig:real-perception}
\vspace{-2mm}
\end{figure}

\begin{table}[!htbp]
\centering
\caption{\small Real-world articulated object manipulation results. We run 10 trials for each task and count the number of successful/half-successful/failed trials respectively, where half-successful trials include behaviors like detaching during pulling and pushing forcefully.}
\label{tab:real-manipulation}
\setlength{\tabcolsep}{5.5pt}
\begin{tabular}{cccccc}
\toprule
\multicolumn{2}{c}{Tasks} & \makecell{PointNet++\\\cite{pointnet++}} & \makecell{ANCSH\\\cite{ancsh}} & \makecell{GAMMA\\\cite{gamma}} & \makecell{RPMArt\\(ours)} \\
\midrule
\multirow{2}{*}{Microwave} & Pull & 6/2/2 & 4/0/6 & 8/1/1 & \textbf{9/1/0} \\
 & Push & 5/4/1 & 3/4/3 & 6/3/1 & \textbf{7/1/2} \\
\hline
\multirow{2}{*}{Refrigerator} & Pull & 2/1/7 & 1/1/8 & 3/1/6 & \textbf{7/0/3} \\
 & Push & 0/0/10 & 1/0/9 & 2/0/8 & \textbf{8/1/1} \\
\hline
\multirow{2}{*}{Safe} & Pull & \textbf{7/0/3} & 5/2/3 & 5/1/4 & \textbf{7/0/3} \\
 & Push & 7/0/3 & \textbf{7/1/2} & \textbf{7/1/2} & \textbf{7/1/2} \\
\hline
\multirow{2}{*}{\makecell{Storage\\Furniture}}& Pull & 1/0/9 & 3/1/6 & 2/1/7 & \textbf{4/0/6} \\
 & Push & 2/2/6 & \textbf{6/2/2} & 2/3/5 & 5/2/3 \\
\hline
\multirow{2}{*}{Drawer} & Pull & 1/1/8 & 2/1/7 & 0/2/8 & \textbf{2/2/6} \\
 & Push & 2/0/8 & 2/1/7 & 0/0/10 & \textbf{3/2/5} \\
 \hline
\multirow{2}{*}{\makecell{Washing\\Machine}} & Pull & 0/0/10 & 0/1/9 & 0/0/10 & \textbf{3/3/4} \\
 & Push & 0/0/10 & 0/0/10 & 0/0/10 & \textbf{1/2/7} \\
\bottomrule
\end{tabular}

\end{table}

\begin{figure}[!htbp]
\centering
\includegraphics[width=\columnwidth]{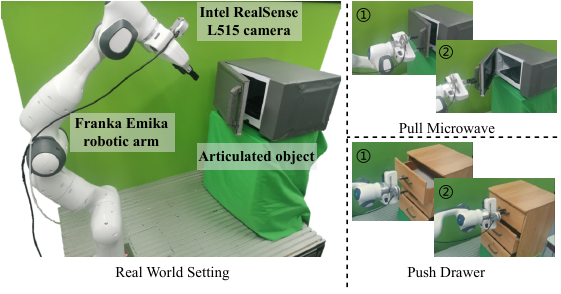}
\caption{\small Real-world manipulation experiments.}
\label{fig:real-manipulation}
\vspace{-5mm}
\end{figure}

To validate the ability for sim-to-real transfer of our framework, we also conduct real-world experiments using our model, trained only on synthetic data.

{\bf Articulation perception.}
We first collect point clouds of six real articulated objects under different conditions, including scenarios with and without background, as well as presence or absence of distractors. We capture depth images for each object with 5 uniformly selected joint states, each from 20 randomly selected camera views. Then we use our trained models to estimate the joint parameters and affordable points. The quantitative results, excluding backgrounds and distractors, are shown in Table \ref{tab:real-perception}. We also visualize the estimation results with both background and distractors included in Fig. \ref{fig:real-perception}. We can find that RoArtNet demonstrates more stable performance compared to other baselines. However, some performance degradation is found in the StorageFurniture and Drawer categories for RoArtNet, as well as for ANCSH and GAMMA. This could possibly be attributed to the relatively small size of parts in these two objects, where all three models somewhat rely on part segmentation to complete the estimation. Another noteworthy observation pertains to the performance on WashingMachine. Specifically, only RoArtNet successfully estimates targets accurately, while the other three baselines exhibit significantly large estimation errors. We find a potential reason that we take a relatively small washing machine toy as the object, then the influence of noisy points is relative significant.

{\bf Articulated object manipulation.}
We also apply the models to manipulate the real articulated objects. We run 10 trials for each task, and count the number of successful, half-successful and failed trials. Here, half-successful trials include behaviors like detaching during pulling and pushing forcefully. Table \ref{tab:real-manipulation} shows the statistics, and Fig. \ref{fig:real-manipulation} illustrates the manipulation process. Videos are available on our website. Our method outperforms other counterparts, especially for Refrigerator and WashingMachine. Refrigerator has a glossy surface, while WashingMachine is relatively small, which makes the noise more prominent.

\section{Conclusion}

We present RPMArt, a framework towards robust perception and manipulation for articulated objects. At its core, RoArtNet learns local context features from sampled point tuples to vote the joint parameters and affordable points robustly. To further improve its capability for sim-to-real transfer, articulation awareness is introduced to account for the unique geometric structure of articulated objects. Finally, we use the estimated affordable point to select the affordable initial grasp pose and generate manipulation actions guided by the estimated joint constraints. Experiments show that RPMArt achieves state-of-the-art performance in both noise-added simulation and real-world environments. Currently, RoArtNet can only achieve category-level generalization. In future work, we will explore methods that can also accomplish robust cross-category estimation.

\section*{Acknowledgements}

This work was supported by the National Key Research and Development Project of China (No. 2022ZD0160102, No. 2021ZD0110704), Shanghai Artificial Intelligence Laboratory, XPLORER PRIZE grants, National Natural Science Foundation of China (No. 52305030, No. 62302143), and Anhui Provincial Natural Science Foundation (No. 2308085QF207).

{\small
\bibliographystyle{IEEEtran}
\bibliography{ref}
}

\end{document}